\documentclass[11pt,a4paper]{article}
\usepackage[hyperref]{clib_acl}

\usepackage{booktabs}
\usepackage{tabularx}
\usepackage{array}
\newcolumntype{Y}{>{\raggedright\arraybackslash}X}
\setcitestyle{notesep={: }}
\usepackage{balance}
\usepackage{times}
\usepackage{latexsym}
\usepackage{graphics,graphicx}

\usepackage[title]{appendix}

\usepackage{microtype}

\aclfinalcopy 

\title{SimpCue: Cue-Based Prompting for Multilingual Text Simplification}

\author{
  Mehrzad Tareh \quad Horacio Saggion \quad Stefan Bott \\
  TALN Group, Universitat Pompeu Fabra (UPF) \\
  Barcelona, Spain \\
  \texttt{\{mehrzad.tareh, horacio.saggion, stefan.bott\}@upf.edu}
}
  
\date{}

\begin{document} 
\maketitle
\begin{abstract}
Text simplification aims to make complex texts easier to understand while preserving their original meaning. Recent large language models can perform simplification through prompting, but it remains unclear whether adding explicit linguistic information about sentence complexity to the prompt improves their outputs. We investigate this question for multilingual sentence-level Easy-to-Read simplification in Catalan, Spanish, and Italian. Using Qwen3-8B, we compare a baseline prompt, a gold-cue prompt enriched with gold linguistic cues, and a predicted-cue prompt enriched with automatically predicted cues. We evaluate the outputs using SARI, BLEU, chrF, and BERTScore, and complement this evaluation with a manual qualitative analysis. Predicted-cue prompting obtains the best overall scores across all four metrics, although the gains over the baseline are small. Gold-cue prompting does not consistently improve over the baseline, and results vary across languages. These findings indicate that cue-based prompting can influence multilingual Easy-to-Read simplification, but its benefits are modest, metric-dependent, and language-dependent.

\end{abstract}

\section{Introduction}

Text simplification (TS) aims to reduce a text’s lexical, syntactic, and discourse complexity while preserving its core meaning, relevant information, and grammatical correctness 
\citep{saggion-2017-automatic-text-simplification, siddharthan_2006_cohesion}. At the sentence level, simplification may involve operations such as lexical substitution, sentence splitting, reordering, deletion, explicitation of implicit information, referent clarification, simplification of numerical expressions, or replacement of complex discourse connectors \citep{siddharthan_katsos_2010_connectives}. This task is especially important for improving cognitive and linguistic accessibility, since administrative, health, educational, civic, and other public information texts often need to be made easier to understand for readers with diverse literacy, language-learning, or cognitive accessibility needs.

Recent Large Language Models (LLMs) have made prompt-based TS increasingly practical, as they can generate fluent simplified text without fine-tuning for the specific task \citep{dinc_etal_2025_itu}. However, a general instruction such as \textit{``simplify this sentence''} does not explicitly identify why the input sentence is difficult. A sentence may be complex because it contains technical vocabulary, abstract terminology, long syntactic dependencies, passive or impersonal constructions, nominalizations, unclear pronoun reference, dense numerical expressions, or several ideas compressed into one sentence \citep{shardlow_etal_2022_predicting, bautista_etal_2013_numerical}. When these sources of complexity are not made explicit, the model must infer them implicitly during generation. As a result, it may address only some comprehension barriers while leaving others unresolved \citep{liu_lee_2025_enhancing}.

This paper investigates whether prompt-based multilingual Easy-to-Read (E2R) simplification can be steered toward explicitly specified simplification criteria by adding linguistic cues to the prompt. We use E2R to refer to text simplification aimed at making information more accessible for readers who may have difficulties processing complex written language \citep{nomura-etal-2010-guidelines, yaneva-etal-2016-evaluating}. In this paper, we use E2R requirements to denote practical simplification principles specified in E2R guidelines and targeting specific linguistic phenomena.

We study this question using Catalan, Spanish, and Italian data from the multilingual text simplification dataset \citep{bott_etal_2026_multilingual_idem_corpus}: the iDEM corpus\footnote{Data available through the MER-TRANS shared task: \url{https://lastus-taln-upf.github.io/mertrans-iberlef-2026/}.}. This dataset contains complex source sentences and human E2R simplifications produced by E2R experts on the basis of recommendations closely aligned with E2R principles. It also includes annotation criteria, complex parts of the source sentence, and alignments between original and simplified text. In this work, we use normalized expert annotation criteria as gold cue labels. Although the same prompt template is used conceptually across languages, it is translated into Catalan, Spanish, or Italian to match the corresponding input language. More broadly, multilingual NLP work has explored language-aware strategies, multilingual models, and prompt engineering to address variation across diverse linguistic settings \citep{mohammadzadeh-etal-2025-iasbs}. These criteria differ in formulation but overlap in their goal of reducing processing difficulty. 
We evaluate three prompting conditions:
\begin{itemize}
    \item \textbf{Baseline}: no additional cues are provided.
    \item \textbf{Gold-cue}: cues are derived from expert annotation criteria.
    \item \textbf{Predicted-cue}: cues are automatically inferred by the proposed method.
\end{itemize}
This comparison separates two related questions: whether explicit linguistic guidance is beneficial for simplification \citep{martin_etal_2020_controllable}, and whether automatically predicted guidance is sufficiently accurate for practical use. If gold cues improve simplification but predicted cues do not, the main limitation may lie in automatic cue prediction rather than in the cue-based prompting strategy itself. Conversely, if neither condition improves over the baseline, this would call into question the practical value of adding explicit cues to simplification prompts. We evaluate cue-based prompting with Qwen3-8B
\citep{yang_etal_2025_qwen3,qwen_2025_qwen3_8b}
for Catalan, Spanish, and Italian E2R simplification.
We select Qwen3-8B because it provides open weights and documented
multilingual support including Catalan, Spanish, and Italian, and a
feasible model size for controlled server-side experiments.
Generated outputs are compared against human E2R references using
automatic metrics including SARI, BLEU, chrF, and BERTScore
\citep{alva_manchego_etal_2019_easse}.
Because we evaluate only Qwen3-8B, our conclusions are model-specific
and should not be interpreted as showing that it is the best model for
E2R simplification in general.
The contributions of this paper are threefold:
\begin{itemize}
    \item \textbf{First}, we evaluate a cue-based prompting strategy for multilingual E2R text simplification.
    \item \textbf{Second}, we conduct a controlled comparison of baseline, gold-cue, and predicted-cue prompting for Catalan, Spanish, and Italian using Qwen3-8B.
    \item \textbf{Third}, we evaluate whether automatically predicted linguistic cues can serve as a practical substitute for expert-derived cues when gold annotations are unavailable.
\end{itemize}
We apply the same cue-based prompting paradigm across all three
languages. However, the specific cues differ across languages due to
the characteristics of the underlying dataset. This somewhat limits
direct cross-lingual comparability, but reflects the available
annotation structure.

\section{Related Work}

\subsection{Text Simplification and Easy-to-Read Texts}

Text simplification is commonly formulated as a rewriting task aimed at making an input text easier to read while preserving its meaning and grammaticality. At the sentence level, simplification may involve lexical substitution, deletion, sentence splitting, reordering, paraphrasing, and other rewriting operations \citep{alva_manchego_etal_2020_data}. Early approaches relied on hand-written rules, lexical resources, statistical machine translation, or tree-based rewriting \citep{chandrasekar_etal_1996_motivations, wubben_etal_2012_sentence}. Later work increasingly adopted neural sequence-to-sequence models trained on aligned complex--simple sentence pairs \citep{nisioi_etal_2017_exploring}, while large-scale resources from Wikipedia and Simple English Wikipedia played an important role in the field's development \citep{coster_kauchak_2011_simple}. E2R simplification is closely related to TS, although E2R guidelines are primarily intended for human authors and translators. Compared with general TS, E2R is more explicitly oriented toward cognitive accessibility and toward producing texts that are understandable for readers with intellectual or cognitive disabilities and other reading difficulties \citep{saggion_2024_ai_nlp_easy_to_read}. This makes E2R particularly relevant to public, legal, institutional, educational, and health information \citep{martinez_etal_2024_exploring_etr_llms}. Most TS resources and benchmarks have historically concentrated on English, while resources for other languages remain comparatively limited \citep{ryan_etal_2023_revisiting}. In multilingual simplification, language is also an important consideration, since lexical, syntactic, and discourse properties may affect both the sources of complexity and the simplification strategies appropriate for a given language. Among Romance languages, Spanish has received attention through the development of simplification resources and systems \citep{bott_saggion_2014_spanish_resources, saggion_2015_simplext}, whereas Catalan and Italian remain comparatively less represented in E2R-oriented resources. More broadly, multilingual NLP research has explored cross-lingual and language-aware modeling strategies for handling variation across linguistic settings \citep{mohammadzadeh-etal-2025-iasbs}. More recent resources, such as the multilingual text simplification corpus, provide original texts and E2R versions produced by experts for Catalan, Spanish, and Italian, enabling comparative evaluation across these languages \citep{bott_etal_2026_multilingual_idem_corpus}. Previous work has also incorporated human-authored E2R guidelines directly into prompts for LLM-based text simplification \citep{brotoortega-2026-translating}. However, to our knowledge, the effect of explicit linguistic cues has not been systematically examined under both gold and automatically predicted cue settings across multiple languages.

\subsection{Controllable Text Simplification}

A closely related line of work investigates controllable simplification. 
Instead of generating a generic simplified output, controllable systems 
condition generation on explicit variables that influence the type or degree 
of simplification. ACCESS introduced control tokens for properties such as 
output length, paraphrasing, lexical complexity, and syntactic complexity 
\citep{martin_etal_2020_controllable}. Building on this approach, 
\citep{sheang-saggion-2021-controllable} incorporated similar control tokens into a 
T5-based simplification model and introduced an additional word-ratio 
control to regulate word length. Other work has explicitly modeled 
simplification operations such as splitting, deletion, and paraphrasing 
\citep{maddela_etal_2021_controllable}, while more recent approaches have 
explored instance-level prediction of the operations required for individual 
inputs \citep{agrawal_carpuat_2023_controlling}. Our work follows this 
input-sensitive motivation but uses a different form of control information: 
linguistic cues derived from expert complexity annotations. These cues 
indicate the specific sources of difficulty identified in each sentence and 
are used to provide more targeted guidance for simplification.

\subsection{Prompt-Based Simplification with LLMs}

Instruction-tuned LLMs have enabled text simplification through prompting without additional fine-tuning. Recent work has explored zero-shot prompting for readability-controlled simplification \citep{farajidizaji-etal-2024-possible, barayan-etal-2025-analysing}, prompt-based approaches using lightweight LLMs \citep{sanchez-gomez-etal-2025-hulat}, and automatic prompt induction and optimization \citep{chernodub-etal-2025-apio}. These studies demonstrate the potential of prompting for text simplification, but general simplification instructions and target readability levels do not explicitly identify which linguistic sources of complexity should be addressed for a particular input. Our proposed approach, SimpCue, addresses this limitation by conditioning the LLM on instance-specific linguistic cues derived from expert complexity annotations and provided either as gold cues or automatically predicted cues. Rather than controlling simplification only through a general instruction or desired readability level, these cues provide linguistically grounded guidance about the specific sources of difficulty to address in each instance.

\section{Dataset}

We evaluate our approach on the iDEM Corpus, a multilingual E2R dataset for
Catalan, Spanish, and Italian
\citep{bott_etal_2026_multilingual_idem_corpus}. It contains complex source
texts and corresponding human-produced E2R versions from public and
institutional domains. It is especially suitable for our study because it provides original and simplified sentence alignments together with expert annotation criteria and expert-created simplifications, which we normalize and use as gold linguistic cue labels.

In this work, each instance includes an original sentence, a human E2R reference, and zero or more linguistic cue labels. The original sentence is treated as the complex input, and the aligned E2R sentence or sentences are treated as the human reference. The cue labels encode simplification-relevant criteria related to lexical difficulty, technical terminology, sentence structure, use of lists, complex verbal forms, pronoun reference, passive or impersonal constructions, nominalizations, difficult connectors, and other language-specific E2R conventions. The dataset is distributed in separate files for Catalan, Spanish, and Italian. We process and evaluate the three languages separately because their annotation schemes, surface cues of complexity, and linguistic realization of simplification operations differ, while using the same underlying model and prompt design across languages. The raw data contains both document-level and sentence-level information, as shown in Table~\ref{tab:dataset_statistics}. For each language, it includes document identifiers, sentence identifiers, language labels, original source texts, human E2R texts, segmented original sentences, segmented simplified sentences, aligned simplified sentence identifiers, annotation criteria, complex-span information where available, and alignments between original and simplified sentences. These alignments are important because simplification is not always one-to-one: one complex sentence may correspond to several simplified sentences, and some simplification operations involve splitting, reordering, deletion, or explicitation of implicit information. 

\begin{table}[t]
\centering
\small
\setlength{\tabcolsep}{3pt}
\renewcommand{\arraystretch}{1.2}
\begin{tabular*}{\columnwidth}{@{\extracolsep{\fill}}lcccccc@{}}
\hline
\textbf{Language} & \textbf{Docs} & \textbf{Original} & \textbf{E2R} & \textbf{Train} & \textbf{Dev} & \textbf{Test} \\
\hline
Catalan & 14 & 383 & 1,380 & 145 & 130 & 108 \\
Spanish & 17 & 354 & 1,287 & 275 & 33 & 46 \\
Italian & 16 & 325 & 718 & 258 & 27 & 40 \\
\hline
Total & 47 & 1,062 & 3,385 & 678 & 190 & 194 \\
\hline
\end{tabular*}
\caption{Dataset statistics after preprocessing, including the number of documents, original sentences, E2R sentence units, and training, development, and test instances for each language.}
\label{tab:dataset_statistics}
\end{table}

\section{Methodology}

\subsection{Task Formulation}

We formulate the task as sentence-level E2R text simplification. Given an original sentence $x_i$ in language $l \in \{\mathrm{CAT}, \mathrm{ES}, \mathrm{IT}\}$, the system generates a simplified output $\hat{y}_i$ in the same language. The output may consist of one sentence or a short sequence of sentences, since E2R rewriting may split one complex sentence into several shorter ones.
Each instance has a human E2R reference $y_i$, which is used only for evaluation. We distinguish between two types of linguistic cues. Gold cues, denoted $g_i$, are derived from the dataset's expert annotation criteria. Predicted cues, denoted $\hat{g}_i$, are produced automatically by language-specific cue-prediction baselines fitted on the training split and selected using the development split.
The three prompting conditions are defined as follows:
\[
\begin{array}{rcl}
\hat{y}^{\mathrm{base}}_{i} &=& M(P_{\mathrm{base}}(x_i)), \\
\hat{y}^{\mathrm{gold}}_{i} &=& M(P_{\mathrm{gold}}(x_i, g_i)), \\
\hat{y}^{\mathrm{pred}}_{i} &=& M(P_{\mathrm{pred}}(x_i, \hat{g}_i)).
\end{array}
\]
Here, $M$ is \texttt{Qwen3-8B}, $P_{\mathrm{base}}$ is the baseline prompt, while $P_{\mathrm{gold}}$ and $P_{\mathrm{pred}}$ are the prompts augmented with gold and predicted cues, respectively.

\subsection{Dataset Preprocessing and Representation}

We convert the raw data into a canonical sentence-level representation that preserves the document identifier, text identifier, language, original sentence identifier, original sentence, human E2R reference, aligned simplified-sentence identifiers, cue annotations, and complex-span information where available. This representation allows us to apply the same experimental logic across Catalan, Spanish, and Italian while preserving language-specific annotation information. Because simplification may involve one-to-many alignments, the pipeline distinguishes between an alignment-level view and a sentence-level view. The alignment-level view is used to inspect how human simplifications split or restructure source sentences. The sentence-level view is used for prompt-based generation and cue prediction, where each source sentence is associated with its gold cues and reconstructed human E2R reference. When one original sentence is aligned to multiple E2R segments, the simplified segments are concatenated in their original order to form a single reference. This preserves sentence splitting as part of the reference while allowing automatic metrics to compare one generated output with one reference field \citep{chamovitz_abend_2022_cognitive}.

\subsection{Gold Cue Mapping and Representation}

Gold cue annotations consist of language-specific criterion codes and their textual descriptions. During preprocessing, we map each code to its standardized description using separate dictionaries for Catalan, Spanish, and Italian. As shown in Table~\ref{tab:cue-inventories}, the complete inventories contain 38, 34, and 33 criteria, respectively, of which 23, 33, and 30 occur in the processed dataset. We preserve these language-specific inventories rather than mapping them to a shared cross-lingual label set. Numerical criterion codes are used as class labels for classification and evaluation, while prompts contain only the corresponding textual descriptions without numerical codes.

\begin{table}[t]
\centering
\small
\setlength{\tabcolsep}{5pt}
\renewcommand{\arraystretch}{1.15}
\begin{tabular}{lccc}
\hline
\textbf{Lang.} & \textbf{Labels (O/C)} & \textbf{Predictor} & \textbf{Selection Criterion} \\
\hline
CAT & 23/38 & LR + Top-$K$ & Dev. micro-F1 \\
ES  & 33/34 & LR           & Dev. micro-F1 \\
IT  & 30/33 & LR + Top-$K$ & Dev. micro-F1 \\
\hline
\end{tabular}
\caption{Cue-label inventories and prediction configurations selected using TF–IDF features. O/C denotes the numbers of observed and candidate labels, respectively; LR denotes logistic regression.}
\label{tab:cue_predictor_selection}
\end{table}

\subsection{Automatic Cue Prediction}

The sizes of the language-specific cue inventories and the selected
cue-prediction configurations are summarized in
Table~\ref{tab:cue_predictor_selection}.
The predicted-cue condition requires an automatic method for assigning cue labels to unseen original sentences. We treat this as a multi-label classification task, since a sentence may require zero, one, or several simplification operations \citep{tsoumakas_katakis_2007_multilabel}. Representative examples of the resulting gold and predicted cue sets are shown in Table ~\ref{tab:representative-cue-classification}. For the predicted-cue condition, we first evaluated several leakage-safe cue prediction baselines on sentence-level train, development, and test splits. We trained cue predictors only on the training split. Model selection was performed independently for each language using development set micro-F$_1$, and test scores were computed only after development-based selection. This procedure ensures that the test split is used only for final reporting. During development, we tested simple supervised models, including frequency-based prediction and
TF--IDF representations \cite{salton-buckley-1988-term} with one-vs-rest
linear classifiers \cite{rifkin-klautau-2004-defense}. More broadly, hybrid NLP approaches combining transformer-based models with classical machine learning classifiers have also been explored for classification in multilingual and code-mixed settings \citep{tareh-2025-pabsa, tareh-etal-2024-iasbs}. We also explored a two-stage approach in which we first generated candidate cue labels and then filtered them with a binary reranker. However, this approach did not outperform the selected simple baselines on development data. We selected the final predicted-cue configuration separately for each language based on development-set micro-F$_1$. For all three languages, predicted cue labels were generated using a TF-IDF one-vs-rest logistic regression model. This configuration was selected after comparison with k-nearest neighbors, support vector machines, DistilBERT, multilingual BERT, and count-vectorizer variants, as it achieved the best development-set performance. For Catalan and Italian, we additionally applied a top-$k$ label-selection strategy, since it improved development performance for these languages. In this strategy, the model assigns the k highest-scoring labels to each instance, where k is determined from the training split only by rounding the mean number of gold cue labels per training instance. Catalan uses k=2, based on a mean training label cardinality of 2.228, and Italian uses k=7, based on a mean training label cardinality of 6.864. For Spanish, the corresponding value was k=3, based on a mean training label cardinality of 3.295; however, the top-k strategy did not improve development performance and was therefore not used. The selected predicted labels were then used to construct predicted-cue prompts for development and test instances only.

\subsection{Prompting Conditions}

Tables~\ref{tab:baseline-prompts} and~\ref{tab:cue-prompts}
summarize the three prompting conditions used in our experiments:
baseline, gold-cue, and predicted-cue. In the baseline condition, the
model receives only the original sentence and a general simplification
instruction. In the gold-cue condition, the model also receives
gold-standard cue annotations from the dataset. In the predicted-cue condition, the model receives the cues predicted by our language-specific classifier.
This design allows us to evaluate the effect of explicit linguistic
cues while avoiding leakage from the human E2R reference. It also
preserves the methodological distinction between gold cues, which test
whether explicit linguistic guidance is useful in principle, and
predicted cues, which test whether such guidance can be used in a more
realistic setting in which expert labels are unavailable. All
conditions use the same source sentence, the same generation model,
and comparable generation settings. The only intended difference
between conditions is whether the prompt contains no cues, gold cues,
or predicted cues.
We used the same prompt for each language, but we translated it separately to Catalan, Spanish, and Italian. This avoids providing simplification
instructions in a language different from that of the input and allows
each prompt to reflect the target language's E2R terminology. The model is instructed to preserve the meaning of the original sentence, use simple vocabulary, and prefer short sentences. Similar simplification instructions have been used in previous LLM-based sentence simplification work \citep{barayan-etal-2025-analysing}. The model is additionally instructed to rewrite the input in the same language and return only the simplified output, without translations, explanations, comments, or metadata. These restrictions are important because the outputs are evaluated automatically.

\begin{table*}[t]
\centering
\scriptsize
\setlength{\tabcolsep}{5pt}
\renewcommand{\arraystretch}{1.05}

\begin{tabularx}{\textwidth}{@{}YYY@{}}
\hline
\textbf{Catalan} & \textbf{Spanish} & \textbf{Italian} \\
\hline

\begin{minipage}[t]{\linewidth}
\vspace{0.2em}
\textbf{Tasca:} reescriu el text original en català fàcil de llegir.

Conserva el significat principal i no afegeixis informació nova.
No afegeixis informació que no estigui justificada pel text original.
Simplifica'l amb estil de lectura fàcil.
Fes servir paraules clares, frases senzilles i una estructura directa.
Si és útil, pots dividir una frase complexa en diverses frases més curtes.
Mantén la resposta en català.
Escriu només el text simplificat.
No expliquis la resposta.
No mencionis etiquetes, cues ni criteris en la sortida.
No copiïs aquest enunciat ni les instruccions.

\medskip
\textbf{Text original:}
A efectes de còmputs de terminis, la recepció de documents en dia inhàbil es considera efectuada el primer dia hàbil següent.

\medskip
\textbf{Text simplificat:}
Per a calcular terminis, si es rep un document en un dia inhàbil, es considera que s'ha rebut el primer dia hàbil que ve.

\vspace{0.2em}
\end{minipage}
&
\begin{minipage}[t]{\linewidth}
\vspace{0.2em}
\textbf{Tarea:} reescribe el texto original en español fácil de leer.

Conserva el significado principal y no añadas información nueva.
No añadas información que no esté respaldada por el texto original.
Simplifícalo con estilo de lectura fácil.
Usa palabras claras, frases sencillas y una estructura directa.
Si es útil, puedes dividir una oración compleja en varias oraciones más cortas.
Mantén la respuesta en español.
Escribe solo el texto simplificado.
No expliques la respuesta.
No menciones etiquetas, cues ni criterios en la salida.
No copies este enunciado ni las instrucciones.

\medskip
\textbf{Texto original:}
Estas limitaciones del poder de las mayorías favorecen la seguridad económica, el clima de negocios, y la interdicción de arbitrariedad de los poderes públicos que hace posible la integración y la economía de mercado.

\medskip
\textbf{Texto simplificado:}
Las limitaciones al poder de las mayorías ayudan a garantizar la seguridad económica, un buen clima para los negocios y la prevención de actos arbitrarios por parte del gobierno. Esto permite que se desarrolle la integración y la economía de mercado.

\vspace{0.2em}
\end{minipage}
&
\begin{minipage}[t]{\linewidth}
\vspace{0.2em}
\textbf{Compito:} riscrivi il testo originale in italiano facile da leggere.

Mantieni il significato principale e non aggiungere nuove informazioni.
Non aggiungere informazioni non supportate dal testo originale.
Semplificalo in uno stile facile da leggere.
Usa parole chiare, frasi semplici e una struttura diretta.
Se è utile, puoi dividere una frase complessa in più frasi brevi.
Mantieni la risposta in italiano.
Scrivi solo il testo semplificato.
Non spiegare la risposta.
Non citare etichette, cue o criteri nella risposta.
Non copiare questa consegna o le istruzioni.

\medskip
\textbf{Testo originale:}
Secondo Miani servirebbe un progetto più ambizioso per incentivare la coltivazione di verdure e ortaggi, che hanno bisogno di meno fertilizzanti, e l'immissione di piante acquatiche che possono produrre ossigeno invece che consumarlo.

\medskip
\textbf{Testo semplificato:}
Secondo Miani, serve un progetto più ambizioso per promuovere la coltivazione di verdure e ortaggi, che richiedono meno fertilizzanti. Si dovrebbe anche introdurre piante acquatiche che producono ossigeno, invece di consumarlo.

\vspace{0.2em}
\end{minipage}
\\
\hline
\end{tabularx}

\caption{Representative language-specific baseline prompts and their corresponding model outputs used in the experiments.}
\label{tab:baseline-prompts}
\end{table*}

\begin{table*}[t]
\centering
\scriptsize
\setlength{\tabcolsep}{5pt}
\renewcommand{\arraystretch}{1.05}

\begin{tabularx}{\textwidth}{@{}l X X X@{}}
\toprule
\textbf{Lang.} &
\textbf{Source sentence ($x_i$)} &
\textbf{Gold cue set ($g_i$)} &
\textbf{Predicted cue set ($\hat{g}_i$)} \\
\midrule

CAT &
A efectes de còmputs de terminis, la recepció de documents en dia
inhàbil es considera efectuada el primer dia hàbil següent. &
$\{2, 8, 18\}$ &
$\{2, 18\}$ \\

ES &
Estas limitaciones del poder de las mayorías favorecen la seguridad
económica, el clima de negocios, y la interdicción de arbitrariedad de
los poderes públicos que hace posible la integración y la economía de
mercado. &
$\{2, 11, 13, 23\}$ &
$\{2, 5, 11, 15, 23, 25\}$ \\

IT &
Secondo Miani servirebbe un progetto più ambizioso per incentivare la
coltivazione di verdure e ortaggi, che hanno bisogno di meno
fertilizzanti, e l'immissione di piante acquatiche che possono produrre
ossigeno invece che consumarlo. &
$\{2, 4, 5, 7, 8, 10, 11, 12, 19, 22, 33\}$ &
$\{2, 4, 5, 7, 8, 10, 11, 20, 22, 23, 26, 33\}$ \\

\bottomrule
\end{tabularx}

\caption{Representative examples from the cue-classification stage,
showing the source sentence $x_i$, the gold cue set $g_i$, and the
predicted cue set $\hat{g}_i$ for each instance.}
\label{tab:representative-cue-classification}
\end{table*}

\begin{table*}[t]
\centering
\scriptsize
\setlength{\tabcolsep}{5pt}
\renewcommand{\arraystretch}{1.05}

\begin{tabularx}{\textwidth}{@{}YYY@{}}
\hline
\textbf{Catalan} & \textbf{Spanish} & \textbf{Italian} \\
\hline

\begin{minipage}[t]{\linewidth}
\vspace{0.2em}
\textbf{Tasca:} reescriu el text original en català fàcil de llegir.

Conserva el significat principal i no afegeixis informació nova.
No afegeixis informació que no estigui justificada pel text original.
Simplifica'l amb estil de lectura fàcil.
Fes servir paraules clares, frases senzilles i una estructura directa.
Si és útil, pots dividir una frase complexa en diverses frases més curtes.
Tingues en compte les cues lingüístiques següents durant la simplificació.
Les cues són orientacions internes: no les mencionis en la resposta final.
Mantén la resposta en català.
Escriu només el text simplificat.
No expliquis la resposta.
No copiïs aquest enunciat ni les instruccions.

\medskip
\textbf{Cues lingüístiques:}

- Tractament directe de l'interlocutor \\
- Paraules freqüents i comprensibles

\medskip
\textbf{Text original:}
A efectes de còmputs de terminis, la recepció de documents en dia inhàbil es considera efectuada el primer dia hàbil següent.

\medskip
\textbf{Text simplificat:}
Quan es calculen terminis, si es rep un document en un dia que no és laborable, es considera que s'ha rebut el primer dia laborable que ve.

\vspace{0.2em}
\end{minipage}
&
\begin{minipage}[t]{\linewidth}
\vspace{0.2em}
\textbf{Tarea:} reescribe el texto original en español fácil de leer.

Conserva el significado principal y no añadas información nueva.
No añadas información que no esté respaldada por el texto original.
Simplifícalo con estilo de lectura fácil.
Usa palabras claras, frases sencillas y una estructura directa.
Si es útil, puedes dividir una oración compleja en varias oraciones más cortas.
Ten en cuenta las cues lingüísticas siguientes durante la simplificación.
Las cues son orientaciones internas: no las menciones en la respuesta final.
Mantén la respuesta en español.
Escribe solo el texto simplificado.
No expliques la respuesta.
No copies este enunciado ni las instrucciones.

\medskip
\textbf{Cues lingüísticas:}

- Punto aparte o salto de línea entre ideas \\
- Uso de verbos frecuentes \\
- Sustantivos cortos, sencillos y habituales \\
- Evitar términos abstractos, extranjerismos y tecnicismos \\
- Explicar o eliminar abreviaturas, siglas y acrónimos \\
- Orden lógico sujeto--verbo--predicado \\
- Añadir sujeto para evitar elipsis 

\medskip
\textbf{Texto original:}
Estas limitaciones del poder de las mayorías favorecen la seguridad económica, el clima de negocios, y la interdicción de arbitrariedad de los poderes públicos que hace posible la integración y la economía de mercado.

\medskip
\textbf{Texto simplificado:}
Estas limitaciones protegen la seguridad económica.
Favorecen el clima de negocios.
Impiden que los poderes públicos actúen de forma arbitraria.
Esto permite la integración y el funcionamiento de la economía de mercado.

\vspace{0.2em}
\end{minipage}
&
\begin{minipage}[t]{\linewidth}
\vspace{0.2em}
\textbf{Compito:} riscrivi il testo originale in italiano facile da leggere.

Mantieni il significato principale e non aggiungere nuove informazioni.
Non aggiungere informazioni non supportate dal testo originale.
Semplificalo in uno stile facile da leggere.
Usa parole chiare, frasi semplici e una struttura diretta.
Se è utile, puoi dividere una frase complessa in più frasi brevi.
Tieni conto delle cue linguistiche seguenti durante la semplificazione.
Le cue sono indicazioni interne: non citarle nella risposta finale.
Mantieni la risposta in italiano.
Scrivi solo il testo semplificato.
Non spiegare la risposta.
Non copiare questa consegna o le istruzioni.

\medskip
\textbf{Cue linguistiche:}

- Punto o virgola per separare più idee \\
- Due punti per introdurre elenchi \\
- Verbi di uso corrente \\
- Avverbi di uso corrente \\
- Aggettivi di uso corrente \\
- Sostantivi brevi, semplici e comuni \\ 
- Evitare termini astratti, forestierismi e tecnicismi \\
- Eliminare il gerundio \\
- Eliminare subordinate relative \\
- Eliminare impersonali e forme passive \\
- Eliminare nominalizzazioni \\
- Elenchi puntati

\medskip
\textbf{Testo originale:}
Secondo Miani servirebbe un progetto più ambizioso per incentivare la coltivazione di verdure e ortaggi, che hanno bisogno di meno fertilizzanti, e l'immissione di piante acquatiche che possono produrre ossigeno invece che consumarlo.

\medskip
\textbf{Testo semplificato:}
Secondo Miani, serve un progetto più grande per aiutare a coltivare verdure e ortaggi. Questi alimenti richiedono meno fertilizzanti. Si potrebbe anche piantare piante che producono ossigeno, invece di consumarlo.

\vspace{0.2em}
\end{minipage}
\\
\hline
\end{tabularx}

\caption{Representative prompts with cue information for each language
and corresponding model outputs. The gold-cue and predicted-cue
conditions use the same instruction template and differ only in the
cue set provided to the model.}
\label{tab:cue-prompts}
\end{table*}

\subsection{Evaluation}

Automatic evaluation of text simplification is challenging because a good simplification should be simpler than the source, fluent, and semantically faithful \citep{sulem_etal_2018_BLEU}. We evaluate generated simplifications against the human E2R references using SARI, BLEU, chrF, and BERTScore. SARI is treated as the primary simplification metric because it evaluates how well a system keeps, deletes, and adds text relative to the source sentence and the human reference \citep{xu_etal_2016_optimizing}. BLEU measures word-level n-gram overlap with the reference \citep{papineni_etal_2002_BLEU}, while chrF captures character-level similarity to the reference \citep{popovic_2015_chrF}. BERTScore is used as a contextual similarity metric between the generated output and the human reference \citep{zhang_etal_2020_bertscore}. Together, these metrics provide complementary automatic evidence about simplification edits, surface overlap, character-level similarity, and semantic similarity. However, we interpret them cautiously because valid simplifications may differ from a single reference while preserving meaning and improving readability. For a fair comparison, we compute all results for each condition using complete-case rows where outputs from the baseline, gold-cue, and predicted-cue conditions are available and valid. Although baseline and gold-cue outputs were generated for the full dataset, predicted-cue outputs are available only for development and test instances. Results are reported overall and by language and prompting condition. The comparison between baseline and gold-cue tests whether correct expert cues help in principle. The comparison between baseline and predicted-cue tests whether automatically predicted cues help in a realistic setting where expert annotations are unavailable. Where statistical comparisons are reported, paired bootstrap tests are treated as exploratory.

\section{Results}

Table~\ref{tab:automatic_evaluation} shows the results for the baseline, gold-cue, and predicted-cue conditions on the complete-case test subset. The test evaluation includes only instances for which outputs from all three conditions are available and non-empty.
\begin{table*}[t]
\centering
\small
\renewcommand{\arraystretch}{1.1}
\begin{tabular}{llccccc}
\hline
\textbf{Language} & \textbf{Condition} & \textbf{Rows} & \textbf{BLEU} & \textbf{SARI} & \textbf{chrF} & \textbf{BERTScore} \\
\hline
Catalan & baseline & 97 & 11.905 & 39.278 & \textbf{39.052} & \textbf{0.7898} \\
Catalan & gold-cue & 97 & 12.674 & 39.465 & 38.977 & 0.7872 \\
Catalan & predicted-cue & 97 & \textbf{13.014} & \textbf{40.226} & 38.978 & 0.7894 \\
\hline
Spanish & baseline & 46 & 18.355 & \textbf{41.538} & 49.516 & 0.8243 \\
Spanish & gold-cue & 46 & 19.216 & 39.647 & 50.033 & \textbf{0.8259} \\
Spanish & predicted-cue & 46 & \textbf{19.801} & 39.495 & \textbf{50.808} & 0.8252 \\
\hline
Italian & baseline & 37 & 10.532 & 37.733 & 42.890 & 0.7970 \\
Italian & gold-cue & 37 & 12.672 & 38.057 & \textbf{43.496} & \textbf{0.7986} \\
Italian & predicted-cue & 37 & \textbf{12.887} & \textbf{39.117} & 43.458 & \textbf{0.7986} \\
\hline
Overall & baseline & 180 & 13.415 & 39.538 & 42.451 & 0.8001 \\
Overall & gold-cue & 180 & 14.573 & 39.222 & 42.665 & 0.7994 \\
Overall & predicted-cue & 180 & \textbf{15.047} & \textbf{39.811} & \textbf{42.865} & \textbf{0.8004} \\
\hline
\end{tabular}
\caption{Automatic evaluation results for the complete case test subset.
BLEU and chrF are computed at the corpus level, SARI is averaged across
sentences, and BERTScore reports the mean contextual similarity to the
human E2R reference.}
\label{tab:automatic_evaluation}
\end{table*}
Overall, the predicted-cue condition achieves the highest scores on BLEU, SARI, chrF, and BERTScore on the test subset. Gold-cue prompting improves overall BLEU, SARI, and chrF over the baseline, while BERTScore is slightly lower. BERTScore is also nearly unchanged. This indicates that providing expert-derived cues does not automatically lead to better simplification performance across all metrics. One possible explanation is that the model may already infer some relevant sources of complexity from the input sentence. Another possibility is that the cue labels, although informative, are sometimes too coarse or heterogeneous to determine the exact rewriting choices needed for an optimal E2R output.
The results for each language show that the effect of cue prompting is not uniform. For Catalan, predicted-cue achieves the highest BLEU and SARI scores, while the baseline condition remains slightly higher on chrF and BERTScore. For Spanish, predicted-cue achieves the highest BLEU and chrF, but the baseline condition obtains the highest SARI score. For Italian, predicted-cue obtains the highest BLEU and SARI, while gold-cue is marginally higher on chrF and ties with predicted-cue on BERTScore. These differences demonstrate that the usefulness of cue prompting depends on the language, the metric considered, the quality of predicted cues, and the interaction between cue information and the generation model.

\subsection{Qualitative Analysis}

Since the SARI scores do not usually say much about compliance with E2R criteria, we carried out a manual analysis to see in what respects the simplifications produced with the inclusion of cues improved over the baseline simplifications. We found that the simplifications produced with cues in the prompt were often, but not always, better than the baseline simplifications. Not all cues seemed to be picked up by the LLM and used to improve simplification strategies. This could be expected because some of the cues are easier to operationalize than others. For example, the ``elimination of ellipsis'' is very hard to do in sentence simplification without providing the context. We found that, surprisingly, the instruction to address the reader directly was often reflected in the generated outputs, and impersonal formulations like ``when the request is sent'' in the baseline simplifications are better treated as ``when you send the request'' in the simplification that used cues. We also observed that outputs generated under both the gold-cue and predicted-cue conditions tended to be better segmented, with the input information typically distributed across a larger number of output sentences.

\subsection{Exploratory Bootstrap Analysis}

The paired bootstrap analysis provides additional evidence that the observed differences are metric-dependent. Under this exploratory procedure, gold-cue and predicted-cue outputs show higher BLEU than the baseline condition, and predicted-cue also shows higher BLEU than gold-cue. These differences should be interpreted as metric-specific evidence of increased reference overlap, not as confirmatory evidence that cue prompting generally improves the quality of simplification.
For SARI, the differences between conditions are not clearly distinguishable under the same exploratory procedure. This is important because SARI is the main metric used to evaluate simplification in this study. Therefore, the results should not be interpreted as showing that cue prompting robustly improves E2R simplification. Rather, they show that cues can affect model outputs and can improve some overlap-based metrics, while their impact on simplification-specific quality remains inconclusive.

\subsection{Discussion}

Taken together, the results suggest that automatically predicted cues can be incorporated into a multilingual E2R prompting pipeline without clear degradation and may improve some reference-overlap metrics. This is relevant because predicted cues represent the more realistic setting in which expert annotations are unavailable for new input sentences. Despite prediction noise, the predicted-cue condition obtains the highest overall BLEU, SARI, chrF, and BERTScore scores on the test subset restricted to complete cases.
At the same time, the results do not support a strong claim that cue prompting universally improves E2R simplification. The clearest gains are observed for BLEU and chrF, while SARI improvements are small and not consistently supported by the exploratory bootstrap analysis. BERTScore remains nearly stable across conditions, suggesting that cue prompting does not substantially affect contextual similarity to the human reference. A careful interpretation is therefore that prompting with cues is promising as a mechanism for influencing multilingual E2R generation, but its effectiveness is modest and depends on the metric and language.
These findings also help clarify the role of gold and predicted cues. Gold-cue prompting tests whether expert-derived information is useful in principle, but the results show that correct cue information alone is not sufficient to guarantee better automatic simplification scores. Predicted-cue prompting tests whether automatically inferred cues can be useful in a realistic setting. Its relatively strong BLEU and chrF scores suggest that predicted cues may help the model produce outputs closer to the reference surface form, but the small and uncertain SARI differences indicate that better cue prediction or more operationally explicit cue formulations may be needed to obtain clearer gains in simplification.
Overall, the experiment supports a cautious conclusion: explicit linguistic cues can shape LLM-based E2R simplification outputs, and predicted cues can be used in a practical prompting pipeline, but the current evidence does not show a consistent or robust improvement across all languages and metrics. Human evaluation and qualitative error analysis are needed to determine whether the observed differences in automatic metrics correspond to improvements in readability, adequacy, accessibility, or E2R compliance for target readers.

\section{Conclusion and Future Directions}

This paper examined whether explicit linguistic cues improve prompt-based multilingual E2R text simplification for Catalan, Spanish, and Italian. Using data from the multilingual E2R dataset, we compared a baseline prompt with two prompting conditions that incorporated linguistic cues: gold-cue prompting, which uses expert annotations, and predicted-cue prompting, which uses automatically predicted linguistic cues. This design allowed us to distinguish between the potential value of linguistic guidance when cues are available and its practical usefulness when cues must be predicted for unseen input. The results do not show a consistent advantage of gold-cue prompting over the baseline condition. Predicted-cue prompting achieves the highest pooled scores across all four automatic metrics on the complete test subset, although the differences are small for some metrics and BERTScore remains nearly unchanged across conditions. These findings suggest that predicted cues can influence simplification outputs while largely preserving contextual similarity to the reference. However, the gains are modest and vary across languages, indicating that cue-based prompting is a promising but still limited form of linguistic guidance for multilingual E2R simplification.

Future work should evaluate this approach on larger datasets, additional languages, and broader E2R domains. Further improvements should also address rare label representation and class imbalance in the cue-prediction component, since more accurate cue identification may strengthen the overall prompting pipeline. Finally, automatic evaluation should be complemented with human evaluation to assess readability, adequacy, accessibility, and the conditions under which cue-based guidance leads to beneficial, neutral, or harmful rewriting decisions.

\section{Limitations}

This study has several limitations. The dataset is relatively small, and label distributions are imbalanced, which may bias cue prediction toward frequent labels and limit generalization to underrepresented categories and unseen texts. 

The evaluation relies on automatic metrics and a single reference simplification, without human or target-reader assessment; these metrics may therefore miss aspects such as factual preservation, fluency, adequacy, readability, and accessibility for readers. In addition, all generation experiments use a single open-weight instruction model, so the findings should not be generalized to other LLMs, model sizes, model families, or decoding settings without further validation. 

Finally, the paired bootstrap analysis is exploratory and was conducted under runtime and data constraints. It should therefore be interpreted as an indicator of possible system behavior, rather than as definitive evidence of consistent improvements across conditions.

\section*{Acknowledgments}

We thank the two anonymous reviewers for the insightful comments. This research is possible thanks to
grant PCI2026-177545-1 funded by MICIU/AEI/10.13039/501100011033/CHIST-ERA (EU) project ``Translation is Not Enough:
Plain Language Adaptation of Multilingual Science'' and to funding from the European Union's Horizon Europe research and innovation program under the Grant Agreement No. 101132431 (iDEM: Innovative and Inclusive Democratic Spaces for Deliberation and Participation). Views and opinions expressed are, however, those of the authors only and do not necessarily reflect those of the European Union. Neither the European Union nor the granting authority can be held responsible for them.

\bibliographystyle{clib_acl_natbib}
\bibliography{clib}

@article{siddharthan_2006_cohesion,
  author  = {Siddharthan, Advaith},
  title   = {Syntactic Simplification and Text Cohesion},
  journal = {Research on Language and Computation},
  volume  = {4},
  number  = {1},
  pages   = {77--109},
  year    = {2006},
  doi     = {10.1007/s11168-006-9011-1}
}

@inproceedings{maddela_etal_2021_controllable,
  author    = {Maddela, Mounica and Alva-Manchego, Fernando and Xu, Wei},
  title     = {Controllable Text Simplification with Explicit Paraphrasing},
  booktitle = {Proceedings of the 2021 Conference of the North American Chapter of the Association for Computational Linguistics: Human Language Technologies},
  pages     = {3536--3553},
  year      = {2021},
  address   = {Online},
  publisher = {Association for Computational Linguistics},
  doi       = {10.18653/v1/2021.naacl-main.277},
  url       = {https://aclanthology.org/2021.naacl-main.277/}
}

@inproceedings{chamovitz_abend_2022_cognitive,
  author    = {Chamovitz, Eytan and Abend, Omri},
  title     = {Cognitive Simplification Operations Improve Text Simplification},
  booktitle = {Proceedings of the 26th Conference on Computational Natural Language Learning},
  pages     = {241--265},
  year      = {2022},
  address   = {Abu Dhabi, United Arab Emirates (Hybrid)},
  publisher = {Association for Computational Linguistics},
  doi       = {10.18653/v1/2022.conll-1.17},
  url       = {https://aclanthology.org/2022.conll-1.17/}
}

@inproceedings{bautista_etal_2013_numerical,
  author    = {Bautista, Susana and Herv{\'a}s, Raquel and Gerv{\'a}s, Pablo and Power, Richard and Williams, Sandra},
  title     = {A System for the Simplification of Numerical Expressions at Different Levels of Understandability},
  booktitle = {Proceedings of the Workshop on Natural Language Processing for Improving Textual Accessibility},
  pages     = {39--48},
  year      = {2013},
  address   = {Atlanta, Georgia},
  publisher = {Association for Computational Linguistics},
  url       = {https://aclanthology.org/W13-1505/}
}

@inproceedings{siddharthan_katsos_2010_connectives,
  author    = {Siddharthan, Advaith and Katsos, Napoleon},
  title     = {Reformulating Discourse Connectives for Non-Expert Readers},
  booktitle = {Human Language Technologies: The 2010 Annual Conference of the North American Chapter of the Association for Computational Linguistics},
  pages     = {1002--1010},
  year      = {2010},
  address   = {Los Angeles, California},
  publisher = {Association for Computational Linguistics},
  url       = {https://aclanthology.org/N10-1144/}
}

@inproceedings{barayan-etal-2025-analysing,
    title = "Analysing Zero-Shot Readability-Controlled Sentence Simplification",
    author = "Barayan, Abdullah and Camacho-Collados, Jose and Alva-Manchego, Fernando",
    booktitle = "Proceedings of the 31st International Conference on Computational Linguistics",
    month = jan,
    year = "2025",
    address = "Abu Dhabi, UAE",
    publisher = "Association for Computational Linguistics",
    url = "https://aclanthology.org/2025.coling-main.452/",
    pages = "6762--6781"
}

@inproceedings{dinc_etal_2025_itu,
  author    = {Din{\c{c}}, Kutay Arda and Bekta{\c{s}}, Fatih and Eryi{\u{g}}it, G{\"u}l{\c{s}}en},
  title     = {{ITU} {NLP} at {TSAR} 2025 Shared Task A Three-Stage Prompting Approach for {CEFR}-Oriented Text Simplification},
  booktitle = {Proceedings of the Fourth Workshop on Text Simplification, Accessibility and Readability (TSAR 2025)},
  pages     = {149--154},
  year      = {2025},
  address   = {Suzhou, China},
  publisher = {Association for Computational Linguistics},
  doi       = {10.18653/v1/2025.tsar-1.11},
  url       = {https://aclanthology.org/2025.tsar-1.11/}
}

@inproceedings{liu_lee_2025_enhancing,
  author    = {Liu, Fengkai and Lee, John S. Y.},
  title     = {Enhancing Readability-Controlled Text Modification with Readability Assessment and Target Span Prediction},
  booktitle = {Proceedings of the 14th Joint Conference on Lexical and Computational Semantics (*SEM 2025)},
  pages     = {293--303},
  year      = {2025},
  address   = {Suzhou, China},
  publisher = {Association for Computational Linguistics},
  doi       = {10.18653/v1/2025.starsem-1.23},
  url       = {https://aclanthology.org/2025.starsem-1.23/}
}

@article{alva_manchego_etal_2020_data,
  author  = {Alva-Manchego, Fernando and Scarton, Carolina and Specia, Lucia},
  title   = {Data-Driven Sentence Simplification: Survey and Benchmark},
  journal = {Computational Linguistics},
  volume  = {46},
  number  = {1},
  pages   = {135--187},
  year    = {2020},
  doi     = {10.1162/coli_a_00370},
  url     = {https://aclanthology.org/2020.cl-1.4/}
}

@article{shardlow_etal_2022_predicting,
  author  = {Shardlow, Matthew and Evans, Richard and Zampieri, Marcos},
  title   = {Predicting Lexical Complexity in English Texts: The {Complex} 2.0 Dataset},
  journal = {Language Resources and Evaluation},
  volume  = {56},
  pages   = {1153--1194},
  year    = {2022},
  doi     = {10.1007/s10579-022-09588-2}
}

@misc{bott_etal_2026_multilingual_idem_corpus,
      title={A Multilingual Human Annotated Corpus of Original and Easy-to-Read Texts to Support Access to Democratic Participatory Processes}, 
      author={Stefan Bott and Verena Riegler and Horacio Saggion and Almudena Rascón Alcaina and Nouran Khallaf},
      year={2026},
      eprint={2603.05345},
      archivePrefix={arXiv},
      primaryClass={cs.CL},
      url={https://arxiv.org/abs/2603.05345}, 
}

@inproceedings{martin_etal_2020_controllable,
  author    = {Martin, Louis and de la Clergerie, {\'E}ric and Sagot, Beno{\^i}t and Bordes, Antoine},
  title     = {Controllable Sentence Simplification},
  booktitle = {Proceedings of the Twelfth Language Resources and Evaluation Conference},
  pages     = {4689--4698},
  year      = {2020},
  address   = {Marseille, France},
  publisher = {European Language Resources Association},
  url       = {https://aclanthology.org/2020.lrec-1.577/}
}

@misc{yang_etal_2025_qwen3,
  author        = {Yang, An and Li, Anfeng and Yang, Baosong and Zhang, Beichen and Hui, Binyuan and Zheng, Bo and Yu, Bowen and Gao, Chang and Huang, Chengen and Lv, Chenxu and Zheng, Chujie and Liu, Dayiheng and Zhou, Fan and Huang, Fei and Hu, Feng and Ge, Hao and Wei, Haoran and Lin, Huan and Tang, Jialong and Yang, Jian and Tu, Jianhong and Zhang, Jianwei and Yang, Jianxin and Yang, Jiaxi and Zhou, Jing and Zhou, Jingren and Lin, Junyang and Dang, Kai and Bao, Keqin and Yang, Kexin and Yu, Le and Deng, Lianghao and Li, Mei and Xue, Mingfeng and Li, Mingze and Zhang, Pei and Wang, Peng and Zhu, Qin and Men, Rui and Gao, Ruize and Liu, Shixuan and Luo, Shuang and Li, Tianhao and Tang, Tianyi and Yin, Wenbiao and Ren, Xingzhang and Wang, Xinyu and Zhang, Xinyu and Ren, Xuancheng and Fan, Yang and Su, Yang and Zhang, Yichang and Zhang, Yinger and Wan, Yu and Liu, Yuqiong and Wang, Zekun and Cui, Zeyu and Zhang, Zhenru and Zhou, Zhipeng and Qiu, Zihan},
  title         = {Qwen3 Technical Report},
  year          = {2025},
  eprint        = {2505.09388},
  archivePrefix = {arXiv},
  primaryClass  = {cs.CL},
  doi           = {10.48550/arXiv.2505.09388}
}

@misc{qwen_2025_qwen3_8b,
  author       = {{Qwen Team}},
  title        = {{Qwen3-8B}},
  year         = {2025},
  howpublished = {Hugging Face model repository},
  url          = {https://huggingface.co/Qwen/Qwen3-8B},
  note         = {Accessed: 2026-06-27}
}

@inproceedings{papineni_etal_2002_bleu,
  author    = {Papineni, Kishore and Roukos, Salim and Ward, Todd and Zhu, Wei-Jing},
  title     = {{BLEU}: A Method for Automatic Evaluation of Machine Translation},
  booktitle = {Proceedings of the 40th Annual Meeting of the Association for Computational Linguistics},
  pages     = {311--318},
  year      = {2002},
  month     = jul,
  address   = {Philadelphia, Pennsylvania, USA},
  publisher = {Association for Computational Linguistics},
  doi       = {10.3115/1073083.1073135},
  url       = {https://aclanthology.org/P02-1040/}
}

@article{xu_etal_2016_optimizing,
  author  = {Xu, Wei and Napoles, Courtney and Pavlick, Ellie and Chen, Quanze and Callison-Burch, Chris},
  title   = {Optimizing Statistical Machine Translation for Text Simplification},
  journal = {Transactions of the Association for Computational Linguistics},
  volume  = {4},
  pages   = {401--415},
  year    = {2016},
  doi     = {10.1162/tacl_a_00107},
  url     = {https://aclanthology.org/Q16-1029/}
}

@inproceedings{zhang_etal_2020_bertscore,
  author    = {Zhang, Tianyi and Kishore, Varsha and Wu, Felix and Weinberger, Kilian Q. and Artzi, Yoav},
  title     = {{BERTScore}: Evaluating Text Generation with {BERT}},
  booktitle = {Proceedings of the 8th International Conference on Learning Representations},
  year      = {2020},
  url       = {https://openreview.net/forum?id=SkeHuCVFDr}
}

@inproceedings{nisioi_etal_2017_exploring,
  author    = {Nisioi, Sergiu and {\v{S}}tajner, Sanja and Ponzetto, Simone Paolo and Dinu, Liviu P.},
  title     = {Exploring Neural Text Simplification Models},
  booktitle = {Proceedings of the 55th Annual Meeting of the Association for Computational Linguistics (Volume 2: Short Papers)},
  pages     = {85--91},
  year      = {2017},
  month     = jul,
  address   = {Vancouver, Canada},
  publisher = {Association for Computational Linguistics},
  doi       = {10.18653/v1/P17-2014},
  url       = {https://aclanthology.org/P17-2014/}
}

@inproceedings{chandrasekar_etal_1996_motivations,
  author    = {Chandrasekar, R. and Doran, Christine and Srinivas, B.},
  title     = {Motivations and Methods for Text Simplification},
  booktitle = {COLING 1996 Volume 2: The 16th International Conference on Computational Linguistics},
  year      = {1996},
  url       = {https://aclanthology.org/C96-2183/}
}

@inproceedings{coster_kauchak_2011_simple,
  author    = {Coster, William and Kauchak, David},
  title     = {Simple English Wikipedia: A New Text Simplification Task},
  booktitle = {Proceedings of the 49th Annual Meeting of the Association for Computational Linguistics: Human Language Technologies},
  pages     = {665--669},
  year      = {2011},
  month     = jun,
  address   = {Portland, Oregon, USA},
  publisher = {Association for Computational Linguistics},
  url       = {https://aclanthology.org/P11-2117/}
}

@inproceedings{wubben_etal_2012_sentence,
  author    = {Wubben, Sander and van den Bosch, Antal and Krahmer, Emiel},
  title     = {Sentence Simplification by Monolingual Machine Translation},
  booktitle = {Proceedings of the 50th Annual Meeting of the Association for Computational Linguistics (Volume 1: Long Papers)},
  pages     = {1015--1024},
  year      = {2012},
  month     = jul,
  address   = {Jeju Island, Korea},
  publisher = {Association for Computational Linguistics},
  url       = {https://aclanthology.org/P12-1107/}
}

@article{saggion_2024_ai_nlp_easy_to_read,
  author  = {Saggion, Horacio},
  title   = {Artificial Intelligence and Natural Language Processing for Easy-to-Read Texts},
  journal = {Revista de Llengua i Dret / Journal of Language and Law},
  volume  = {82},
  pages   = {84--103},
  year    = {2024},
  doi     = {10.58992/rld.i82.2024.4362},
  url     = {https://doi.org/10.58992/rld.i82.2024.4362}
}

@article{martinez_etal_2024_exploring_etr_llms,
  author  = {Mart{\'i}nez, Paloma and Ramos, Alberto and Moreno, Lourdes},
  title   = {Exploring {Large Language Models} to generate {Easy to Read} content},
  journal = {Frontiers in Computer Science},
  volume  = {6},
  pages   = {1394705},
  year    = {2024},
  doi     = {10.3389/fcomp.2024.1394705},
  url     = {https://doi.org/10.3389/fcomp.2024.1394705}
}

@inproceedings{ryan_etal_2023_revisiting,
  author    = {Ryan, Michael J. and Naous, Tarek and Xu, Wei},
  title     = {Revisiting non-{E}nglish Text Simplification: A Unified Multilingual Benchmark},
  booktitle = {Proceedings of the 61st Annual Meeting of the Association for Computational Linguistics (Volume 1: Long Papers)},
  pages     = {4898--4927},
  year      = {2023},
  month     = jul,
  address   = {Toronto, Canada},
  publisher = {Association for Computational Linguistics},
  doi       = {10.18653/v1/2023.acl-long.269},
  url       = {https://aclanthology.org/2023.acl-long.269/}
}

@article{bott_saggion_2014_spanish_resources,
  author  = {Bott, Stefan and Saggion, Horacio},
  title   = {Text Simplification Resources for {S}panish},
  journal = {Language Resources and Evaluation},
  volume  = {48},
  number  = {1},
  pages   = {93--120},
  year    = {2014},
  doi     = {10.1007/s10579-014-9265-4},
  url     = {https://doi.org/10.1007/s10579-014-9265-4}
}

@article{saggion_2015_simplext,
  author  = {Horacio Saggion and Sanja {\v{S}}tajner and Stefan Bott and
             Simon Mille and Luz Rello and Biljana Drndarevic},
  title   = {Making It Simplext: Implementation and Evaluation of a Text
             Simplification System for Spanish},
  journal = {ACM Transactions on Accessible Computing},
  volume  = {6},
  number  = {4},
  pages   = {14:1--14:36},
  year    = {2015},
  doi     = {10.1145/2738046}
}

@inproceedings{agrawal_carpuat_2023_controlling,
  author    = {Agrawal, Sweta and Carpuat, Marine},
  title     = {Controlling Pre-trained Language Models for Grade-Specific Text Simplification},
  booktitle = {Proceedings of the 2023 Conference on Empirical Methods in Natural Language Processing},
  pages     = {12807--12819},
  year      = {2023},
  month     = dec,
  address   = {Singapore},
  publisher = {Association for Computational Linguistics},
  doi       = {10.18653/v1/2023.emnlp-main.790},
  url       = {https://aclanthology.org/2023.emnlp-main.790/}
}

@inproceedings{sulem_etal_2018_bleu,
  author    = {Sulem, Elior and Abend, Omri and Rappoport, Ari},
  title     = {{BLEU} is Not Suitable for the Evaluation of Text Simplification},
  booktitle = {Proceedings of the 2018 Conference on Empirical Methods in Natural Language Processing},
  pages     = {738--744},
  year      = {2018},
  month     = oct # {-} # nov,
  address   = {Brussels, Belgium},
  publisher = {Association for Computational Linguistics},
  doi       = {10.18653/v1/D18-1081},
  url       = {https://aclanthology.org/D18-1081/}
}

@article{tsoumakas_katakis_2007_multilabel,
  author  = {Tsoumakas, Grigorios and Katakis, Ioannis},
  title   = {Multi-Label Classification: An Overview},
  journal = {International Journal of Data Warehousing and Mining},
  volume  = {3},
  number  = {3},
  pages   = {1--13},
  year    = {2007},
  doi     = {10.4018/jdwm.2007070101}
}

@inproceedings{alva_manchego_etal_2019_easse,
    title = "{EASSE}: {E}asier Automatic Sentence Simplification Evaluation",
    author = "Alva-Manchego, Fernando  and
      Martin, Louis  and
      Scarton, Carolina  and
      Specia, Lucia",
    booktitle = "Proceedings of the 2019 Conference on Empirical Methods in Natural Language Processing and the 9th International Joint Conference on Natural Language Processing (EMNLP-IJCNLP): System Demonstrations",
    month = nov,
    year = "2019",
    address = "Hong Kong, China",
    publisher = "Association for Computational Linguistics",
    url = "https://aclanthology.org/D19-3009",
    doi = "10.18653/v1/D19-3009",
    pages = "49--54",
}

@inproceedings{popovic_2015_chrf,
    title = "chr{F}: character n-gram {F}-score for automatic {MT} evaluation",
    author = "Popovi{\'c}, Maja",
    booktitle = "Proceedings of the Tenth Workshop on Statistical Machine Translation",
    month = sep,
    year = "2015",
    address = "Lisbon, Portugal",
    publisher = "Association for Computational Linguistics",
    url = "https://aclanthology.org/W15-3049/",
    doi = "10.18653/v1/W15-3049",
    pages = "392--395"
}

@book{saggion-2017-automatic-text-simplification,
  author    = {Saggion, Horacio},
  title     = {Automatic Text Simplification},
  series    = {Synthesis Lectures on Human Language Technologies},
  publisher = {Morgan \& Claypool Publishers},
  year      = {2017},
  address   = {San Rafael, CA},
  pages     = {1--137},
  doi       = {10.2200/S00700ED1V01Y201602HLT032},
  isbn      = {978-1-62705-868-1}
}

@manual{nomura-etal-2010-guidelines,
  title        = {Guidelines for Easy-to-Read Materials},
  author       = {Nomura, Misako and Nielsen, Gyda Skat and Tronbacke, Bror},
  year         = {2010},
  organization = {International Federation of Library Associations and Institutions},
  edition      = {2nd}
}

@inproceedings{yaneva-etal-2016-evaluating,
  title     = {Evaluating the Readability of Text Simplification Output for Readers with Cognitive Disabilities},
  author    = {Yaneva, Victoria and Temnikova, Irina and Mitkov, Ruslan},
  booktitle = {Proceedings of the Tenth International Conference on Language Resources and Evaluation (LREC'16)},
  pages     = {293--299},
  year      = {2016},
  address   = {Portoro{\v{z}}, Slovenia},
  publisher = {European Language Resources Association}
}

@inproceedings{farajidizaji-etal-2024-possible,
    title = "Is It Possible to Modify Text to a Target Readability Level? An Initial Investigation Using Zero-Shot Large Language Models",
    author = "Farajidizaji, Asma and Raina, Vatsal and Gales, Mark",
    booktitle = "Proceedings of the 2024 Joint International Conference on Computational Linguistics, Language Resources and Evaluation (LREC-COLING 2024)",
    month = may,
    year = "2024",
    address = "Torino, Italia",
    publisher = "ELRA and ICCL",
    url = "https://aclanthology.org/2024.lrec-main.815/",
    pages = "9325--9339"
}

@inproceedings{sanchez-gomez-etal-2025-hulat,
    title = "{HULAT}-{UC}3{M} at {TSAR} 2025 Shared Task: A Prompt-Based Approach using Lightweight Language Models for Readability-Controlled Text Simplification",
    author = "Sanchez-Gomez, Jesus M. and Moreno, Lourdes and Mart{\'i}nez, Paloma and Sanchez-Escudero, Marco Antonio",
    booktitle = "Proceedings of the Fourth Workshop on Text Simplification, Accessibility and Readability (TSAR 2025)",
    month = nov,
    year = "2025",
    address = "Suzhou, China",
    publisher = "Association for Computational Linguistics",
    url = "https://aclanthology.org/2025.tsar-1.15/",
    doi = "10.18653/v1/2025.tsar-1.15",
    pages = "183--192"
}

@inproceedings{chernodub-etal-2025-apio,
    title = "{APIO}: Automatic Prompt Induction and Optimization for Grammatical Error Correction and Text Simplification",
    author = "Chernodub, Artem and Saini, Aman and Huh, Yejin and Kulkarni, Vivek and Raheja, Vipul",
    booktitle = "Proceedings of the 15th International Conference on Recent Advances in Natural Language Processing - Natural Language Processing in the Generative AI Era",
    month = sep,
    year = "2025",
    address = "Varna, Bulgaria",
    publisher = "INCOMA Ltd., Shoumen, Bulgaria",
    url = "https://aclanthology.org/2025.ranlp-1.28/",
    pages = "234--239"
}

@inproceedings{mohammadzadeh-etal-2025-iasbs,
    title = "{IASBS} at {S}em{E}val-2025 Task 11: Ensembling Transformers for Bridging the Gap in Text-Based Emotion Detection",
    author = "Tareh, Mehrzad  and
      Mohammadzadeh, Erfan  and
      Mohandesi, Aydin  and
      Ansari, Ebrahim",
    booktitle = "Proceedings of the 19th International Workshop on Semantic Evaluation (SemEval-2025)",
    month = jul,
    year = "2025",
    address = "Vienna, Austria",
    publisher = "Association for Computational Linguistics",
    url = "https://aclanthology.org/2025.semeval-1.96/",
    pages = "695--702",
    ISBN = "979-8-89176-273-2",
}

@inproceedings{tareh-etal-2024-iasbs,
    title = "{IASBS} at {S}em{E}val-2024 Task 10: Delving into Emotion Discovery and Reasoning in Code-Mixed Conversations",
    author = "Tareh, Mehrzad  and
      Mohandesi, Aydin  and
      Ansari, Ebrahim",
    booktitle = "Proceedings of the 18th International Workshop on Semantic Evaluation (SemEval-2024)",
    month = jun,
    year = "2024",
    address = "Mexico City, Mexico",
    publisher = "Association for Computational Linguistics",
    url = "https://aclanthology.org/2024.semeval-1.179/",
    doi = "10.18653/v1/2024.semeval-1.179",
    pages = "1229--1238",
}

@misc{tareh-2025-pabsa,
      title={PABSA: Hybrid Framework for Persian Aspect-Based Sentiment Analysis}, 
      author={Mehrzad Tareh and Aydin Mohandesi and Ebrahim Ansari},
      year={2025},
      eprint={2510.04291},
      archivePrefix={arXiv},
      primaryClass={cs.CL},
      url={https://arxiv.org/abs/2510.04291}, 
}

@inproceedings{sheang-saggion-2021-controllable,
    title = "Controllable Sentence Simplification with a Unified Text-to-Text Transfer Transformer",
    author = "Sheang, Kim Cheng  and
      Saggion, Horacio",
    booktitle = "Proceedings of the 14th International Conference on Natural Language Generation",
    month = aug,
    year = "2021",
    address = "Aberdeen, Scotland, UK",
    publisher = "Association for Computational Linguistics",
    url = "https://aclanthology.org/2021.inlg-1.38/",
    doi = "10.18653/v1/2021.inlg-1.38",
    pages = "341--352",
}

@article{salton-buckley-1988-term,
  title={Term-weighting approaches in automatic text retrieval},
  author={Salton, Gerard and Buckley, Christopher},
  journal={Information Processing \& Management},
  volume={24},
  number={5},
  pages={513--523},
  year={1988},
  doi={10.1016/0306-4573(88)90021-0}
}

@article{rifkin-klautau-2004-defense,
  title={In Defense of One-Vs-All Classification},
  author={Rifkin, Ryan and Klautau, Aldebaro},
  journal={Journal of Machine Learning Research},
  volume={5},
  pages={101--141},
  year={2004}
}

@inproceedings{brotoortega-2026-translating,
    title = {Translating {Easy}-to-{Read} {Standards} into {Prompts}: {An} {Empirical} {Study} of {LLM}-{Based} {Text} {Simplification}},
    author = {Broto-Ortegal, Juan and Francisco, Virginia and Hervás, Raquel},
    booktitle = {Computers {Helping} {People} with {Special} {Needs}: 20th {International} {Conference}, {ICCHP} 2026, {Brno}, {Czech} {Republic}, {July} 15–17, 2026, {Proceedings}, {Part} {I}},
    publisher = {Springer Nature},
    pages = {411–419},
    doi       = {10.1007/978-3-032-31285-3_49},
    volume    = {16866},
    year      = {2026},
}

\begin{appendices}

\section{Additional Results for Cue Prediction}
Monolingual training performs slightly better than combined CAT+ES+IT training for all three languages, as shown in Table~\ref{tab:cue-prediction-results}. We therefore use separate cue prediction models for each language in the final pipeline. Adding a top-k label prior based on training frequency improves performance for Catalan and Italian, but not for Spanish.

\begin{table*}[t]
\centering
\small
\setlength{\tabcolsep}{6pt}
\renewcommand{\arraystretch}{1.05}

\begin{tabular*}{\textwidth}{@{\extracolsep{\fill}}lcccc@{}}
\hline
& \multicolumn{2}{c}{\textbf{Training setting}} &
\multicolumn{2}{c}{\textbf{Top-$k$ comparison}} \\
\cline{2-3}
\cline{4-5}
\textbf{Language} &
\textbf{Combined} &
\textbf{Monolingual} &
\textbf{BC} &
\textbf{BC + top-$k$ prior} \\
\hline
Catalan & 0.68 & \textbf{0.79} & 0.79 & \textbf{0.83} \\
Spanish & 0.48 & \textbf{0.51} & \textbf{0.51} & 0.47 \\
Italian & 0.69 & \textbf{0.72} & 0.72 & \textbf{0.77} \\
\hline
\end{tabular*}

\caption{Micro F1 results for cue prediction. BC denotes TF--IDF
with logistic regression, with and without a top $k$ prior.}
\label{tab:cue-prediction-results}
\end{table*}

\section{Language-Specific Cue Inventories}

The three languages use separate cue inventories derived from their
language-specific annotation criteria. These inventories are not assumed
to be cross-lingually equivalent or one-to-one aligned.
Table~\ref{tab:cue-inventories} reports all candidate cue codes and
their compact descriptions.

\begin{table*}[]
\centering
\small
\setlength{\tabcolsep}{2pt}
\renewcommand{\arraystretch}{1.03}

\begin{minipage}[t]{0.325\textwidth}
\textbf{Catalan (38 cues)}
\vspace{0.25em}

\begin{tabularx}{\linewidth}{@{}rX@{}}
\hline
\textbf{ID} & \textbf{Cue} \\
\hline
1  & Estructura dels paràgrafs \\
2  & Tractament directe de l'interlocutor \\
3  & Informació contextual inicial \\
4  & Llenguatge no discriminatori \\
5  & Estructura clara dels continguts \\
6  & Títols informatius \\
7  & Limitar el contingut a l'essencial \\
8  & Glosses i glossaris \\
9  & Evitar referències encreuades \\
10 & Llistes i enumeracions \\
11 & Diàlegs literaris \\
12 & Resums \\
13 & Activitats de comprensió \\
14 & Sintaxi simple i frases curtes \\
15 & Prioritzar frases afirmatives \\
16 & Temps verbals simples \\
17 & Prioritzar la veu activa \\
18 & Paraules freqüents i comprensibles \\
19 & Verbs en lloc de nominalitzacions \\
20 & Evitar sentit figurat, ironia i sarcasme \\
21 & Evitar comparatius i superlatius complexos \\
22 & Adverbis freqüents; evitar \textit{-ment} \\
23 & Evitar manlleus i tecnicismes \\
24 & Conjuncions i connectors simples \\
25 & Preposicions habituals \\
26 & Pronoms amb referència clara \\
27 & Modificadors freqüents \\
28 & Evitar sinònims per al mateix referent \\
29 & Evitar formes verbals poc freqüents \\
30 & Evitar ambigüitat per polisèmia \\
31 & Puntuació i caràcters especials \\
32 & Ús de majúscules \\
33 & Nombres i quantitats \\
34 & Números de telèfon \\
35 & Dates i hores \\
36 & Percentatges \\
37 & Números romans \\
38 & Abreviatures, sigles i acrònims \\
\hline
\end{tabularx}
\end{minipage}
\hfill
\begin{minipage}[t]{0.325\textwidth}
\textbf{Spanish (34 cues)}
\vspace{0.25em}

\begin{tabularx}{\linewidth}{@{}rX@{}}
\hline
\textbf{ID} & \textbf{Cue} \\
\hline
1  & Mayúsculas al inicio y en nombres propios \\
2  & Punto aparte o salto de línea entre ideas \\
3  & Evitar signos de puntuación poco habituales \\
4  & Dos puntos para introducir listas \\
5  & Uso de verbos frecuentes \\
6  & Pronombres con referencia clara \\
7  & Uso de adverbios frecuentes \\
8  & Eliminar adverbios terminados en \textit{-mente} \\
9  & Uso de adjetivos frecuentes \\
10 & Eliminar superlativos \\
11 & Sustantivos cortos, sencillos y habituales \\
12 & Evitar términos abstractos, extranjerismos y tecnicismos \\
13 & Definir términos técnicos o complejos necesarios \\
14 & Evitar polisemia o aclararla con contexto \\
15 & Explicar o eliminar abreviaturas, siglas y acrónimos \\
16 & Escribir números pequeños con cifras \\
17 & Simplificar números de muchos dígitos \\
18 & Evitar fracciones, porcentajes y ordinales \\
19 & Indicativo y tiempos verbales sencillos \\
20 & Eliminar tiempos verbales compuestos \\
21 & Eliminar gerundios \\
22 & Eliminar perífrasis verbales complejas \\
23 & Orden lógico sujeto--verbo--predicado \\
24 & Eliminar oraciones impersonales \\
25 & Añadir sujeto para evitar elipsis \\
26 & Eliminar dobles negaciones \\
27 & Eliminar nominalizaciones \\
28 & Eliminar sarcasmo, metáforas y humor \\
29 & Eliminar elipsis \\
30 & Eliminar incisos y aposiciones \\
31 & Simplificar conectores complejos \\
32 & Trato directo al lector \\
33 & Títulos que anticipen el contenido \\
34 & Formato de listado \\
\hline
\end{tabularx}
\end{minipage}
\hfill
\begin{minipage}[t]{0.325\textwidth}
\textbf{Italian (33 cues)}
\vspace{0.25em}

\begin{tabularx}{\linewidth}{@{}rX@{}}
\hline
\textbf{ID} & \textbf{Cue} \\
\hline
1  & Maiuscole a inizio frase e nei nomi propri \\
2  & Punto o virgola per separare più idee \\
3  & Evitare segni di punteggiatura insoliti \\
4  & Due punti per introdurre elenchi \\
5  & Verbi di uso corrente \\
6  & Pronomi con riferimento chiaro \\
7  & Avverbi di uso corrente \\
8  & Aggettivi di uso corrente \\
9  & Eliminare superlativi non correnti \\
10 & Sostantivi brevi, semplici e comuni \\
11 & Evitare termini astratti, forestierismi e tecnicismi \\
12 & Definire tecnicismi o termini complessi necessari \\
13 & Eliminare o chiarire termini polisemici \\
14 & Eliminare o spiegare abbreviazioni e acronimi \\
15 & Scrivere i numeri in cifra \\
16 & Semplificare numeri grandi \\
17 & Evitare frazioni, percentuali e ordinali \\
18 & Indicativo e tempi verbali semplici \\
19 & Eliminare condizionale e congiuntivo \\
20 & Eliminare il gerundio \\
21 & Ordine logico soggetto--verbo--oggetto \\
22 & Eliminare subordinate relative \\
23 & Eliminare impersonali e forme passive \\
24 & Specificare il soggetto \\
25 & Eliminare la doppia negazione \\
26 & Eliminare nominalizzazioni \\
27 & Eliminare sarcasmo, metafore e umorismo \\
28 & Eliminare ellissi \\
29 & Eliminare incisi e apposizioni \\
30 & Eliminare congiunzioni non correnti \\
31 & Rivolgersi direttamente a chi legge \\
32 & Titoli che anticipano il contenuto \\
33 & Elenchi puntati \\
\hline
\end{tabularx}
\end{minipage}

\caption{Candidate cue inventories for Catalan, Spanish, and Italian.
Cue IDs follow the mappings used in the experiments; 23, 33, and 30
cues occur in the processed data, respectively.}
\label{tab:cue-inventories}
\end{table*}

\section{Generation Configuration}

We generated all simplifications using \texttt{Qwen/Qwen3-8B} with
Hugging Face Transformers 5.12.1 and PyTorch 2.5.1 on an NVIDIA
TITAN Xp GPU. We used a batch size of 1, disabled sampling, and
limited generation to 256 new tokens. For the baseline and gold-cue
conditions, we used greedy decoding with a repetition penalty of 1.0. Prompts were formatted using the Qwen chat template and provided as a
single user message. No system message was used, and thinking mode was
disabled.

\end{appendices}
\end{document}